\documentclass[twoside]{article}
\usepackage{aistats2015}

\usepackage{graphicx} 
\usepackage{subfigure} 
\usepackage{amsmath,amsthm}
\usepackage{slashbox}
\usepackage{multirow}
\usepackage{amssymb}

\newcommand{\argmax}{\operatornamewithlimits{argmax}}

\usepackage{hyperref}

\begin{document}

%

%

\twocolumn[

\aistatstitle{Sequential Labeling with online Deep Learning}

\aistatsauthor{ Gang Chen, Ran Xu \and Sargur (Hari) Srihari}
\aistatsaddress{ Dept. of Computer Science and Engineering, SUNY at Buffalo\\
Buffalo, NY 14260\\
\{gangchen,rxu2\}@buffalo.edu, srihari@cedar.buffalo.edu } ]
\begin{abstract}
In this paper, we leverage both deep learning and conditional random fields (CRFs) for sequential labeling. More specifically, we propose a mixture objective function to predict labels either independent or correlated in the sequential patterns. We learn model parameters in a simple but effective way. In particular, we pretrain the deep structure with greedy layer-wise restricted Boltzmann machines (RBMs), followed with an independent label learning step. Finally, we update the whole model with an online learning algorithm, a mixture of perceptron training and stochastic gradient descent to estimate model parameters. We test our model on different challenge tasks, and show that this simple learning algorithm yields the state of the art results.  
 \end{abstract}

\section{Introduction}\label{intro}
Recent advances in deep learning \cite{Hinton06a,Vincent10,Bengio12} have sparked great interest in dimension reduction \cite{Hinton06b,Weston08} and classification problems \cite{Hinton06a,Larochelle12}. In a sense, the success of deep learning lies on learned features, which are useful for supervised/unsupervised tasks \cite{Erhan10,Bengio12}. For example, the binary hidden units in the discriminative Restricted Boltzmann Machines (RBMs) \cite{Larochelle08,Gelfand10} can model latent features of the data to improve classification. 
Unfortunately, one major difficulty in deep learning \cite{Hinton06a} is structured output prediction \cite{Mnih11}, where output space typically may have an exponential number of possible configurations. 
As for sequential labeling, the joint classification of all the items is also difficult because observations are of an indeterminated dimensionality and the number of possible classes is exponentially growing in the length of the sequences.

Fortunately, graphic model is a powerful tool for structure representation, and can address the complex output scenarios effectively. For example, Conditional Random Fields (CRFs) is discriminative probabilistic model for structured prediction \cite{Lafferty01}, which has been widely used in natural language processing \cite{Collins03,Sha03}, handwriting recognition \cite{Taskar03,Maaten11}, speech recognition \cite{Prabhavalkar10} and scene parsing \cite{Kumar03,Shotton06}. One of key advantages of CRFs 
can be attributed to its exploitation on context information and its structured output prediction. 
However, linear CRFs with the raw data input strongly restricts its representation power for classification tasks. More recently, one trend is to generalize CRFs to learn discriminative and non-linear representations, such as 
kernel CRFs \cite{Lafferty04}, hidden-unit CRFs \cite{Peng09,Maaten11} and CRFs with multilayer perceptrons \cite{LeCun98,Prabhavalkar10}. 
As an alternative, some studies have trained CRFs on feature representations learned by unsupervised deep learning \cite{Mohamed09}. 
Hence, how to learn the non-linear features in CRFs is vital to improve classification performance. 

\begin{figure}
\centering
\includegraphics[trim = 20mm 24mm 20mm 30mm, clip, width=8.5cm]{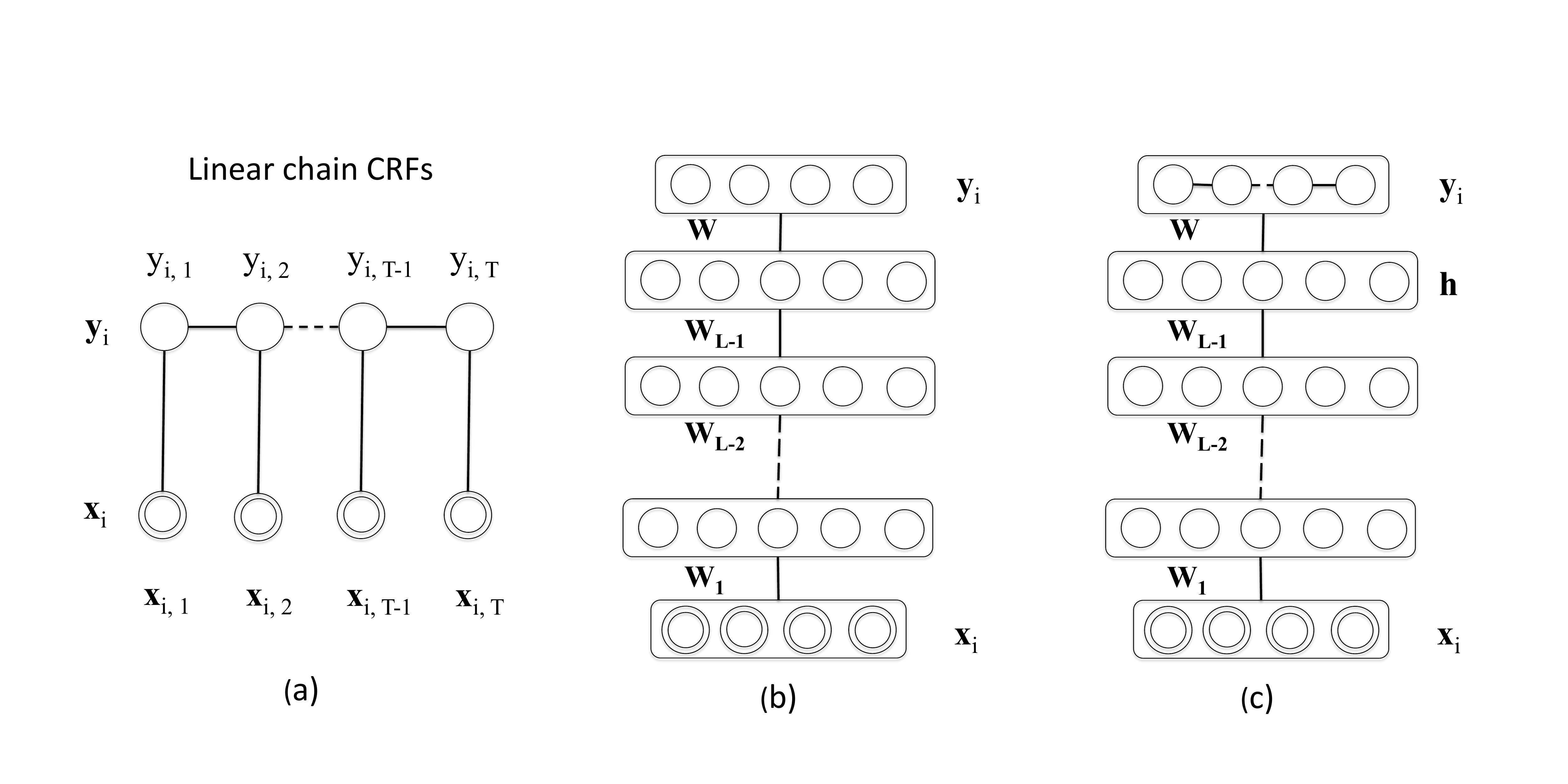}
\caption{(a) linear chain CRFs; (b) deep neural networks (for classification); (c) our deep neural networks for sequential labeling. The two main differences between (b) and (c) are: (1) ${\bf y}_i$ in (c) is a label sequence, which has links between labels, while ${\bf y}_i$ in (b) is a single label with vector representation; (2) the input of (c) is a sequence with multiple instances (or frames), while the input of (b) is an independent instance (or vector). }
\label{fig:fig1}
\end{figure}
On the one hand, target labeling can be better predicted from learned latent representations than raw input data. While deep learning can model the non-linear property of the data and learn representations for better classification, but it cannot effectively handle complex output. On the other hand, when labels exhibit inter-relationships it is imperative to capture context information for structured output prediction. 
Thus, some work has exploited to learn deep non-linear features under linear-chain CRFs \cite{Peng09,Do10}. However, little attention has been paid to handle the overfitting problems because a large number of parameters needs to estimate in the deep neural networks. Moreover, how to effectively pretrain the deep neural network and update the parameters online is still a challenge. 

In this paper, we propose a model for sequential labeling with deep learning, which inherits both advantages of linear chain CRFs and deep learning. Hence, our model can learn non-linear features and also handle structured output, refer to Fig. \ref{fig:fig1} for visual understanding about the model. We pre-train our model with stacked RBMs for feature representations, followed with independently labeling learning under backpropagation. And then we optimize the whole model parameters using an online learning algorithm, which is a mixture of perceptron training and stochastic gradient descent. In particular, we train the top layer with perceptron algorithm, while learning the weights of the lower layers in the deep structure with stochastic gradient descent. Thus, our model is more powerful than linear Conditional Random Fields (CRFs) because the objective function learns latent non-linear features so that target labeling can be better predicted. We test our model over a range of tasks and show that it yields accuracy significantly better than the state of the art.


\section{Sequential labeling with deep learning}
Let $D = \{ \langle {\bf x}_{i}, {\bf y}_{i}\rangle \}_{i=1}^N$ be a set of $N$ training examples. Each example is a pair of a time series $\langle {\bf x}_{i}, {\bf y}_{i}\rangle$, with ${\bf x}_{i} = \{{\bf x}_{i,1}, {\bf x}_{i,2},...,{\bf x}_{i,T_i}\}$ and ${\bf y}_{i} = \{{ y}_{i,1},{y}_{i,2},...,{y}_{i,T_i}\}$, where ${\bf x}_{i,t}  \in \mathbb{R}^d$ is the $i$-th observation at time $t$ and ${y}_{i,t}$ is the corresponding label (we indicate its encoded vector as ${\bf y}_{i,t}$ that uses a so-called 1-of-$K$ encoding). Linear first-order CRFs \cite{Lafferty01} is a conditional discriminative model over the label sequence given the data 
\begin{equation}
\centering
p({\bf y}_{i} | {\bf x}_{i}) = \frac{\textrm{exp}\{  E({\bf x}_{i}, {\bf y}_{i}) \}}{Z({\bf x}_{i})}
\label{eq:eqcrf}
\end{equation}
where $Z({\bf x}_{i})$ is the partition function and $E( {\bf x}_{i}, {\bf y}_{i})$ is the energy function given by
\begin{align}
&E( {\bf x}_{i}, {\bf y}_{i}) =  {\bf y}_{i,1}^T \boldsymbol{\pi} + {\bf y}_{i, T_i}^T \boldsymbol{\tau} \nonumber \\
+ & \sum_{t=1}^{T_i}  ({\bf x}_{i,t}^T {\bf W} {\bf y}_{i,t}  + {\bf b}^T {\bf y}_{i,t} )  + \sum_{t=2}^{T_i}  {\bf y}_{i, t-1}^T {\bf A} {\bf y}_{i,t}
\label{eq:eqe}
\end{align}
where ${\bf y}_{i,1}^T \boldsymbol{\pi}$ and ${\bf y}_{i, T_i}^T \boldsymbol{\tau}$ are the initial-state and final-state factors respectively, ${\bf b}^T {\bf y}_{i,t}$ is the bias term for labels, ${\bf A} \in \mathbb{R}^{K \times K}$ represents the state transition parameters and ${\bf W} \in \mathbb{R}^{d \times K}$ represents the parameters of the data-dependent term.
One of the main disadvantages of linear CRFs is the linear dependence on the raw input data term. Thus, we introduce our sequential labeling model with deep feature learning, which leverages both context information, as well as the nonlinear representations in the deep learning architecture \cite{Hinton06b}. 

\subsection{Objective function}
Although it is possible to leverage the deep neural networks for structured prediction, its output space is explosively growing because of non-determined length of sequential data. Thus, we consider a compromised model, which combine CRFs and deep learning in an unified framework, refer Fig. (\ref{fig:fig1}). 
We propose an objective function with $L$ layers neural network structure, 
\begin{align}
 & \mathcal{L}(D; \boldsymbol{\theta}, \boldsymbol{\omega}) = - \sum_{i=1}^N \textrm{log} p({\bf y}_{i,1},..., {\bf y}_{i,T_i}| {\bf h}_{i,1},...,  {\bf h}_{i,T_i})  \nonumber \\
 + &\frac {\lambda_1}{2} \sum_{i=1}^N \sum_{t=1}^{T_i}  ||\underbrace{f_{L} \circ f_{L-1}\circ \cdot\cdot\cdot \circ f_{1}}_{L\textrm{ times}} ({\bf x}_{i,t}) - {\bf y}_{i,t} ||^2  \nonumber \\
 + & \lambda_2 ||\boldsymbol{ \theta}||^2 + \lambda_3 ||\boldsymbol{\omega}||
\label{eq:eqobj}
\end{align}
where $\boldsymbol{\theta}$ and $\boldsymbol{\omega}$ are the top layer parameters and lower layer ($l = \{1,...,L-1\}$) parameters respectively, which will be explained later. The first row on the right side of the equation is from the linear CRFs  in Eq. (\ref{eq:eqcrf}), but with latent features, which depends respectively on $\boldsymbol{\theta}$ and the latent non-linear features ${\bf h}_i = \{{\bf h}_{i,1},..,{\bf h}_{i,T_i} \}$ in the coding space, with
\begin{align}
& \textrm{log} p({\bf y}_{i,1},..., {\bf y}_{i,T_i}| {\bf h}_{i,1},...,  {\bf h}_{i,T_i})   \nonumber \\
= & \sum_{t=2}^{T_i}  {\bf y}_{i, t-1}^T {\bf A} {\bf y}_{i,t} +   \sum_{t=1}^{T_i}  \big({\bf h}_{i,t}^T {\bf W} {\bf y}_{i,t}  + {\bf b}^T {\bf y}_{i,t}  \big)  \nonumber \\
+ & {\bf y}_{i,1}^T \boldsymbol{\pi} + {\bf y}_{i,T_i}^T \boldsymbol{\tau} - \textrm{log} (Z({\bf h}_{i}))
\label{eq:linearpart}
\end{align}
and non-linear mappings ${\bf h}_{i}$ is the output with $L-1$ layers neural network, s.t. 
\begin{equation}\label{eq:hidden}
{\bf h}_{i} = \underbrace{f_{L-1} \circ f_{L-2}\circ \cdot\cdot\cdot \circ f_{1}}_{L-1 \textrm{ times}}({\bf x}_{i})
\end{equation}
where $\circ$ indicates the function composition, and $f_i$ is logistic function with the weight parameter ${\bf W}_l$ respectively for $l = \{1,..,L-1\}$, refer more details in Sec. \ref{sec:learning}. With a bit abuse of notation, we denote ${\bf h}_{i,t}  = f_{1 \rightarrow (L-1)} ({\bf x}_{i,t})$. 

The second term in the right hand side of Eq. (\ref{eq:eqobj}) is for deep feature learning, with the top layer defined as
 \begin{align}\label{eq:top}
 & \underbrace{f_{L} \circ f_{L-1}\circ \cdot\cdot\cdot \circ f_{1}}_{L\textrm{ times}} ({\bf x}_{i,t}) \nonumber \\
 =& f_{1 \rightarrow L}({\bf x}_{i,t}) =  f_{L}({\bf h}_{i,t}) =  {\bf h}_{i,t}^T {\bf W} + {\bf c}^T {\bf h}_{i,t }
\end{align}
where ${\bf W}$ has been defined in Eq. (\ref{eq:linearpart}), and ${\bf c}$ is the bias term. 
Note that ${\bf W}$ is the same in both Eqs. \ref{eq:linearpart} and \ref{eq:top}. Hence, the second term can be thought as the label prediction independently without considering context information. The weighing ${\lambda_1}$ can control the balance between the first term and the second one on the RHS in Eq. (\ref{eq:eqobj}). If ${\lambda_1} \rightarrow +\infty$, then Eq. (\ref{eq:eqobj}) can be thought as the deep learning \cite{Hinton06b} for classification without context information, and it can handle the cases where outputs are independent (no significant patterns in the label sequences). If ${\lambda_1} \rightarrow 0$, then Eq. (\ref{eq:eqobj}) is the CRFs with non-linear deep feature learning, which generalizes the linear CRFs to learn non-linear deep mappings.

The last two terms in Eq. (\ref{eq:eqobj}) are for regularization on the all parameters
with $\boldsymbol{\theta} = \{ {\bf A}, {\bf W}, \boldsymbol{\pi}, \boldsymbol{\tau}, \boldsymbol{b}, \boldsymbol{c}\}$, and $\boldsymbol{\omega} = \{ {\bf W}_l |  l \in [1,..,L-1] \} $. We add the $\ell_2$ regularization to $\boldsymbol{\theta}$ as most linear CRFs does, while we have the $\ell_1$-regularized term on weight parameters $\boldsymbol{\omega}$ in the deep neural network to avoid overfitting in the learning process.

The aim of our objective function in Eq. (\ref{eq:eqobj}) is for sequential labeling, which explores both the advantages of Markov properties in CRFs and latent representations in deep learning. Our model is different from the common deep learning structure in Fig. \ref{fig:fig1}(b). Firstly, the input to our model in Fig. \ref{fig:fig1}(c) is the sequential data, such as sequences with non-determined length, while the input to Fig. \ref{fig:fig1}(b) is generally an instance with fixed length. Secondly, our model can predict structured outputs or label sequences, while the output in Fig. \ref{fig:fig1}(b) is just one label for each instance, which is independent without context information. Note that we use the first-order CRFs for clarity in Eq. \ref{eq:linearpart} and the rest of the paper, which can be easily extended for the second or high-order cases. Lastly, we use an online algorithm in our deep learning model for parameter updating, which has the potential to handle large scale dataset.

\subsection{Parameter learning}\label{sec:learning}
We use RBMs to initialize the weights layer by layer greedily, with contrast divergence \cite{Hinton06a} (we used CD-1 in our experiments). Then we compute the sub-gradient w.r.t. $\boldsymbol{\theta}$ and $\boldsymbol{\omega}$ in the objective function, and optimize it with online learning. 

{\bf Initialization}: The second term on the right hand side of Eq. (\ref{eq:eqobj}) is from the deep belief network (DBN) for classification in \cite{Hinton06a}. In our deep model, the weights from the layers $1$ to $L-1$ are ${\bf W}_l$ respectively, for $l = \{1,..,L-1\}$, and the top layer $L$ has weight ${\bf W}$. We first pre-train the $L$-layer deep structure with RBMs layer by layer greedily. 
Specifically, we think RBM is a 1-layer DBN, with weight ${\bf W}_1$. Thus, DBN can learn parametric nonlinear mapping from input ${\bf x}$ to output ${\bf h}$, $f: {\bf x} \rightarrow {\bf h}$. For example, for 1-layer DBN, we have ${\bf h} = f_1({\bf x})  = \textrm{logistic}({\bf W_{1}}^T [{\bf x}, 1])$, where we extend ${\bf x} \in \mathbb{R}^d$ into $[{\bf x},1] \in \mathbb{R}^{(d+1)}$ in order to handle bias in the non-linear mapping. 
Note that we use the logistic function from layers $1$ to $L-1$, and the top layer is linear mapping with weight ${\bf W}$ in our deep neural network.

After learned the representation for the data, we minimize 
$ \lambda_1 \sum_{i=1}^N \sum_{t=1}^{T_i}  ||\underbrace{f_{L} \circ f_{L-1}\circ \cdot\cdot\cdot \circ f_{1}}_{L\textrm{ times}} ({\bf x}_{i,t}) - {\bf y}_{i,t} ||$ with L-BFGS (backpropagation is used to compute subgradient w.r.t. weights in each layer) to learn all the weights. More specifically, to learn the initial weights in the deep network, we think each instance ${\bf x}_{i,t} \in {\bf x}_{i}$ has its corresponding label ${\bf y}_{i,t} \in {\bf y}_{i}$ independently. Then, the parameters can be finetuned \cite{Hinton06b}. Note that it does not leverage the context information in this stage, and we will show the independent label learning step is helpful to boost the recognition accuracy. 
Finally, we will update the parameters $\boldsymbol{\theta}$ and $\boldsymbol{\omega}$ in an online way simultaneously, which will be introduced in the following parts. 

{\bf Learning}:
In training the CRFs with deep feature learning, our aim is to minimize objective function $\mathcal{L}(D; \boldsymbol{\theta}, \boldsymbol{\omega}) $ in Eq. (\ref{eq:eqobj}). Because we introduce the deep neural network here for feature learning, the objective is not convex function anymore. However, we can find a local minimum in Eq. (\ref{eq:eqobj}). In our learning framework, we optimize the objective function with an online learn algorithm, by mixing perceptron training and stochastic gradient descent. 

Firstly, we can calculate the (sub)gradients for all parameters. Considering different regularization methods for $\boldsymbol{\theta}$ and $\boldsymbol{\omega}$ respectively, we can calculate gradients w.r.t. them separately. As for the parameters in the negative log likelihood in Eq. \ref{eq:eqobj}, we can compute the gradients w.r.t. $\boldsymbol{\theta}$ as follows
\begin{subequations}
\begin{equation}
\frac{\mathcal{\partial L}}{\partial {\bf A}} = \sum_{i=1}^N \sum_{t=2}^{T_i} {\bf y}_{i,t-1}  ({\bf y}_{i,t})^T - \boldsymbol{\gamma}_{i,t-1}  (\boldsymbol{\gamma}_{i,t})^T;  \label{eq:grad0}
\end{equation}
\begin{equation}
\frac{\mathcal{\partial L}}{\partial \boldsymbol{\pi}} = \sum_{i=1}^N ({\bf y}_{i,1} - \boldsymbol{\gamma}_{i,1});  \label{eq:grad1}
\end{equation}
\begin{equation}
\frac{\partial \mathcal{L}}{\partial \boldsymbol {\tau}} = \sum_{i=1}^N ({\bf y}_{i,T_i} - \boldsymbol{\gamma}_{i,T_i}); \label{eq:grad2} \\
\end{equation}
\begin{equation} 
\frac{\mathcal{\partial L}}{\partial {\bf b}} = \sum_{i=1}^N \big( \sum_{t =1}^{T_i}( {\bf y}_{i,t} -  \boldsymbol{\gamma}_{i,t}) \big); \label{eq:grad3} \\
\end{equation}
\begin{align}
 \frac{\mathcal{\partial L}}{\partial {\bf W}} & = \sum_{i=1}^N \sum_{t =1}^{T_i}  \big( {\bf h}_{i,t} ( {\bf y}_{i,t} - \boldsymbol{\gamma}_{i,t})^T  \nonumber \\
 & +  \lambda_1  f_{1 \rightarrow (L-1)} ({\bf x}_{i,t})  ( {\bf y}_{i,t} -  \hat{\bf y}_{i,t})^T  \big)
 \end{align}
\label{eq:grad}
\end{subequations} 
where $\boldsymbol{\gamma}_{i,t} \in \mathbb{R}^K$ is the vector of length $K$, which can be thought as the posterior probability for labels in the sequence and will be introduced in Sec. \ref{sec:infer}, and $\hat{\bf y}_{i,t} =  f_{1 \rightarrow L}({\bf x}_{i,t})$ is the output from Eq. (\ref{eq:top}). Note that it is easy to derive the gradients of the $\ell_2$ regularization term w.r.t. $\boldsymbol{\theta}$ in the objective in Eq. (\ref{eq:eqobj}), which can be added to the gradients in Eq. (\ref{eq:grad}). 

As for the gradients of weights $\boldsymbol{\omega} = \{ {\bf W}_l |  l \in [1,..,L-1] \} $, we first use backpropagation to get the partial gradient in the neural network, refer to \cite{Hinton06b} for more details. Then the gradient of the $\ell_1$ term in Eq. (\ref{eq:eqobj}) can be attached to get the final gradients w.r.t. ${\bf W}_{l}$ for $l=\{1,..,L-1\}$.

Finally, we use a mixture of perceptron learning and stochastic gradient descent to optimize the objective function. As mentioned in \cite{Maaten11}, they are various optimization methods, such as L-BFGS \cite{Byrd95,Rasmussen05}, stochastic gradient descent (SGD) and perceptron-based learning. L-BFGS as a gradient descent method, has been widely used to optimize weights in the deep structure \cite{Hinton06a}. However, it can be slow, and there are no guarantees if there are multiple local minima in the error surface. SGD and perceptron training both are the online learning algorithms by updating the parameters using the gradient induced by a single time series, so they have significant computational advantages over L-BFGS. Furthermore, perceptron-based online learning can be viewed as a special case of SGD, but it is more flexible than SGD on parameter updating (i.e. parallelization). In our experiments, we tried L-BFGS, but it can be easily trapped into the bad local minimum, and performs worse than other optimization methods in almost all experiments. Thus, in this work, we use perceptron-based learning for the CRF related parameters and stochastic gradient descent for the parameters in the deep structure in all our experiments.

If the perceptron incorrectly classifies a training example, each of the input weights is nudged a little bit in the Òright directionÓ for that training example. In other words, we only need to update the CRF related parameters only for frames that are misclassified in each training example. To update the CRF related parameters with perceptron learning, we need to find the most violated constraints for each example. Basically, given a training example $\langle  {\bf x}_{i}, {\bf y}_{i} \rangle $, we infer its most violated labeling ${\bf y}_i^{\ast}$. If the frame is misclassified, then it directly performs a type of stochastic gradient descent on the energy gap between the observed label sequence and the predicted label sequence. Otherwise, we do not need to update the model parameters. Thus, for the parameters $\boldsymbol{\theta}$ from the negative log likelihood in Eq. (\ref{eq:eqobj}), we first project ${\bf x}_i$ into the code ${\bf h}_{i}$ according to Eq. (\ref{eq:hidden}). Then, the updating rule takes the form below
\begin{equation}
\boldsymbol{\theta} \leftarrow  \boldsymbol{\theta}  + \eta_{\boldsymbol{\theta} }   \frac{\partial }{\partial \boldsymbol {\theta}} \big({E({\bf h}_{i}, {\bf y}_{i}) - E({\bf h}_{i}, {\bf y}_i^{\ast})} \big)
\label{eq:theta}
\end{equation}
where ${\bf y}_i^{\ast}$ is the most violated constraint in the misclassificated case, and $\eta_{\boldsymbol{\theta} }$ is a parameter step size. Note that the posterior probability $\boldsymbol{\gamma}_{i,t} \in \mathbb{R}^K$ in Eq. (\ref{eq:grad}) should be changed into the hard label assignment ${\bf y}_{i,t}^{\ast}$ in the inference stage. 

While for the weights $\boldsymbol{\omega}$ in the deep neural network, we first use backpropagation to compute the gradients, and then update it as follows
\begin{equation}
\boldsymbol{\omega} \leftarrow  \boldsymbol{\omega}  - \eta_{\boldsymbol{\omega} }   \frac{\partial \mathcal{L}}{\partial \boldsymbol {\omega}}
\label{eq:omega}
\end{equation}
where $\eta_{\boldsymbol{\omega} }$ is the step size for the parameters.

\subsection{Inference}\label{sec:infer}
Given the observation ${\bf x}_{i} = \{ {\bf x}_{i,1},...,{\bf x}_{i,T_i} \}$, we first use Eq. (\ref{eq:hidden}) to compute the non-linear code ${\bf h}_{i} = \{ {\bf h}_{i,1},...,{\bf h}_{i,T_i} \}$. To simplify the problem, we assume the first-order CRFs here. To estimate the parameters $\boldsymbol{\theta}$, there are two main inferential problems that need to be solved during learning: (1)  the posterior probability (or the marginal distribution of a label given the codes) $\gamma_{i,t}(k) = p(y_{i,t} = k | {\bf h}_{i,1},...,{\bf h}_{i,T_i})$; (2) the distribution over a label edge $\xi_{i,t}(j,k) = p(y_{i,t} =j, y_{i,t+1} = k | {\bf h}_{i,1},...,{\bf h}_{i,T_i})$. The inference problem can be solved efficiently with Viterbi algorithm \cite{Rabiner89,Bishop06}. 

For the given hidden sequence ${\bf h}_{i} = \{ {\bf h}_{i,1},...,{\bf h}_{i,T_i} \}$, we assume the corresponding states $\{ q_{i,1},...,q_{i,T_i} \}$. Furthermore, we define the forward messages $\alpha_{i,t}(k) \propto p(y_{i,1},..,y_{i,t}, q_{i,t} = k | {\bf h}_{i,1},...,{\bf h}_{i,T_i})$, and the backward messages
$\beta_{i,t}(k) \propto p(y_{i,t+1},..,{y}_{i,T_i} |  q_{i,t} = k, {\bf h}_{i,1},...,{\bf h}_{i,T_i})$
\begin{align}
&\alpha_{i, t+1} (j) = \bigg[ \sum_{k=1}^K\alpha_{i,t}(k) A_{kj} \bigg] B(j, y_{i,t+1}); \\
&\beta_{i,t}(j) = \sum_{k=1}^K A_{jk} B(k, y_{i,t+1}) \beta_{i,t+1}(k);
\end{align}
where $B(k, y_{i,t})$ is the probability to emit $y_{i,t}$ at the state $k$. We can compute it as follows
\begin{equation}
B(:, y_{i,t}) = exp\{ {\bf h}_{i,t}^T {\bf W}   + {\bf b}^T +  \lambda_1 f_{L}({\bf h}_{i,t}) \}
\label{eq:emit}
\end{equation}
After calculate $\alpha_{i, t+1} (j) $ and $\beta_{i,t}(j)$, we can compute the marginal probability for $\gamma_{i,t}$ and $\xi_{i,t}$ respectively
\begin{align}
& \gamma_{i,t}(k) \propto \alpha_{i,t} (k) \beta_{i,t}(k), \\
& \xi_{i,t}(k,j) \propto \alpha_{i,t}(k) A_{kj} B(j, y_{i,t+1}) \beta_{i, t+1}(j);
\label{eq:postprob}
\end{align}
Then, we can compute $\boldsymbol{\gamma}_{i,t}$ in Eq. (\ref{eq:grad}), which is the concatenation: $[\gamma_{i,t}(1),...,\gamma_{i,t}(K)]$.

In the testing stage, the main inferential problem is to compute the most likely label sequence ${\bf y}_{1,...,T}^{\ast}$ given the data ${\bf x}_{1,...,T}$ by $\argmax_{{{\bf y}\prime}_{1,...,T}} p({{\bf y}\prime}_{1,...,T} | {\bf x}_{1,...,T} )$, which can be addressed similarly using the Viterbi algorithm mentioned above.

\section{Experiments} 
To test our method, we compared our method to the state of the art approaches and performed experiments on four tasks: (1) optical character recognition, (2) labeling questions and answers, (3) protein secondary structure prediction, and (4) part-of-speech tagging. Below, we described the datasets we used and also the parameter setting in the experiments.

\subsection{Data sets}
1. The OCR dataset \cite{Taskar03} contains data for 6, 877 handwritten words with 55 unique words, in which each word ${\bf x}_i$ is represented as a series of handwritten characters $\{ {\bf x}_{i1},...,{\bf x}_{i, T_i} \}$. The data consists of a total of 52, 152 characters (i.e., frames), with 26 unique classes. Each character is a binary image of size $16 \times 8$ pixels, leading to a 128-dimensional binary feature vector.  
%

2. The FAQ data set \cite{McCallum00} contains data of 48 files on questions and answers, with a total of 55,480 sentences (i.e., frames). Each sentence is a 24- dimensional binary feature that describes lexical characteristics of the sentence. Each sentence in the FAQ data set belongs to one of four labels: (1) question, (2) answer, (3) header, or (4) footer. 

3. The CB513 contains amino acid structures of 513 proteins \cite{Cuff99}, and has been widely used for protein secondary structure prediction. For each of the proteins, it has 20-dimensional position-specific score matrix features. In the experiment, we concatenate the features from the surrounding 13 frames into the 260 dimensional vector \cite{Maaten11}. As common in protein secondary structure prediction, we convert the eight-class labeling into a three-class labeling. The resulting data set has 513 sequence with total 74, 874 frames (260 dimensions), belongs to 3 classes.

4. The Penn Treebank corpus\footnote{\url{www.cis.upenn.edu/~treebank}} has 74, 029 sentences with a total of 1, 637, 267 words. The whole data set contains 49, 115 unique words, and each word in each sentence is labeled according to its part of speech with total 43 different tags. 
To represent each word, all features are measured in a window with width 3 around the current word, which leads to a total of 212, 610 features. If we use 1000 hidden nodes, then we need to store $2\times 10^8$ parameters in the one-layer neural network. Considering the high storage demanding for the personal computer, we calculated the frequency for each dimension in the total 212, 610 features, and selected the most frequent $5000$ features as our codebook. Then we can represent each word with 5000 dimensions in our experiment.

In our experiments, we used the four data sets in \cite{Maaten11}, which are available on the author's website\footnote{\url{http://cseweb.ucsd.edu/~lvdmaaten/hucrf/Hidden-Unit_Conditional_Random_Fields.html}}.  
\subsection{Experimental Setup}
In our experiments, we randomly initialized the weight ${\bf W}$ by sampling from the normal Gaussian distribution, and all other parameters in $\boldsymbol{\theta}$ to be zero (i.e. biases $\boldsymbol{b}$ and $ \boldsymbol{c}$, and the transition matrix ${\bf A}$ all to be zero). As for $\boldsymbol{\omega} = \{ {\bf W}_l |  l \in [1,..,L-1] \} $, we initialized them with DBN, which had been mentioned before. As for the number of layers and the number of hidden units in each layer, we set differently according to the dimensionality for different datasets. In all the experiments, we use the 3-layer deep autoencoder on the four datasets. Considering the OCR dataset has 128 dimensional binary feature, while FAQ is 24 dimensional vector, we set the number of hidden nodes [100 100 64] in each layer respectively on both the OCR dataset and FAQ dataset. For the CB513 dataset, we set the number of hidden units to be [400 200 100]. For the treebank dataset, the hidden units [1000 400 200] are used in the 3-layer network. We did not try other deep structure in the experiments. 

Unless otherwise indicated, we use the average generalization error to measure all methods in 10-fold cross-validation experiments. 

Because L-BFGS can be easily trapped into local minimum, we used perceptron training \cite{Gelfand10} to estimate the CRF related parameters and stochastic gradient descent to learn the weights in the deep neural network. In all experiments with perceptron learning, we did not use regularization terms. In other words, we set $\lambda_2 =0$. And $\lambda_3= 2\times10^{-4}$ for weights in the deep network. To better leverage the context information, we had a low parameter setting with $\lambda_1 = 0.1$. For each dataset, we divided it into $10$ folds (9 folds as the training set, and the rest as the testing/validation set), and performed 100 full sweeps through the training data, to update the model parameters. We tuned the base step size based on the error on a small held-out validation set. From the base step size, we computed parameter-specific step sizes $\eta_{\boldsymbol{\theta} }$ and $\eta_{\boldsymbol{\omega} }$ as suggested by \cite{Gelfand10}. 

\subsection{Results}
We tested our method on the four data sets mentioned above with the second-order label chains.

\begin{table}
\centering
\resizebox{0.8\columnwidth}{!}{%
\begin{tabular}{ |l|c| }
\hline
 \multicolumn{2}{ |c| }{{\bf Hand-written recognition} (\%)} \\\cline{1-2}
 Linear-chain CRF \cite{Do10} & 14.2 \\
\hline
  Max-margin Markov net \cite{Do10}  & 13.4\\
\hline
Searn \cite{Daume09} & 9.09\\ 
\hline
SVM + CRF \cite{Hoefel08} & 5.76 \\
\hline
Deep learning \cite{Hinton06b} & 4.0 \\
\hline
NeuroCRFs \cite{Do10} & 4.44 \\
\hline
Cond. graphical models  \cite{Perez-Cruz07} & 2.7\\
\hline
LSTM \cite{Graves08} & 2.30\\ 
\hline
Hidden-unit CRF  \cite{Maaten11} & 1.9\\
\hline
Our method (without pretrain)  & 1.56 \\
\hline
Our method  & {\bf 0.63} \\
\hline
\end{tabular}
}
\caption{The experimental comparisons on the OCR dataset. Our method (without pretrain) is from the result without independent label learning step. The results reveal the merits of our method over other methods.}
\label{tab:ocr}
\end{table}

In Table \ref{tab:ocr}, we compared the performance of our method with the performance of competing models on the handwriting recognition task. It shows that the pretraining stage in our model is helpful to improve the recognition accuracy (boosting error rate from 1.56\% to 0.63\%). It also shows that the label correlation is helpful in this case. For example, the deep learning without label correlation yields accuracy 4.0\%, while is significantly lower than our model. Compared to shallow learning methods, our method yields a generalization error of $0.63\%$, while the best performance of other methods is $1.9\%$. It demonstrates that our model is significantly better than other methods, and the deep structure is definitely helpful than the shallow models. 

On the FAQ data set, the lowest generalization error of hidden-unit CRFs is $4.43\%$, compared to 3.34\% for our method in Table \ref{tab:faqs}. And again, our method outperforms other competitive baselines. It also shows that the CRF with deep feature learning (3 layers) in this case, is better than the one hidden layer CRF. Note that we just used the original 24 dimension features in the experiment, instead of extending the feature set into a 24 + 242 = 600-dimensional feature representation in \cite{Maaten11}. 

We also test our method on the protein secondary structure prediction task. The results of these experiments are presented in Table \ref{tab:protein}. In particular, our method achieves a generalization error of only $3.16\%$, compared to $19.5\%$ error with the conditional neural field on the CB513 data set. The results presented in the figure indicate that the CRFs with deep feature learning can significantly improve the performance, compared to hidden-unit CRFs.

Lastly, we also tested our method on part-of-speech tagging task. Note that we already take context information into consideration by using a window width 3 for feature representations. And the final representation is based on only 5,000 codebooks because of storage problem for model parameters. To test whether our method can tackle overfitting problem effectively, we randomly sampled a subset from the Penn Treebank corpus, and did the 10 fold cross validation. We show the experimental results in Table \ref{tab:treebank}. It demonstrates that when there's a few data set available for training, deep learning with L-BFGS has overfitting problems. As the number of training data increasing, the performance of the deep learning also is increasing. While our method outperforms other baselines remarkably, and show stable and better performance with increasing training data. It also shows that our method can generalize well effectively, and it is more robust  with few training data in the recognition task.  

\begin{table}
\centering
\resizebox{0.8\columnwidth}{!}{%
\begin{tabular}{ |l|c| }
\hline
 \multicolumn{2}{ |c| }{{\bf FAQ} (\%)} \\\cline{1-2}
 Linear SVM & 9.87\\
 \hline
 Linear CRF  \cite{Maaten11} & 6.54\\
\hline
NeuroCRFs \cite{Do10} & 6.05 \\
\hline
 Hidden-unit CRF  \cite{Maaten11} & 4.43\\
 \hline
 Deep learning \cite{Hinton06b} & 7.75\\
 \hline
Our method (without pretrain) & 7.44 \\ 
\hline
Our method  & {\bf 3.34} \\
\hline
\end{tabular}
}
\caption{The comparison (generalization errors) on the FAQ dataset using different methods. It shows that our method is significantly better than the Hidden-unit CRF.}
\label{tab:faqs}
\end{table}

\begin{table}
\centering
\resizebox{0.99\columnwidth}{!}{%
\begin{tabular}{ |l|c| }
\hline
 \multicolumn{2}{ |c| }{{\bf Protein secondary structure prediction} (\%)} \\\cline{1-2}
 PSIRED \cite{Jones99} & 24.0 \\
\hline
SVM  \cite{Kim03} & 23.4 \\
\hline
 SPINE\cite{Dor07}  & 23.2\\
\hline
YASSP \cite{Karypis06} & 22.2\\ 
\hline
Cond. neural field  \cite{Peng09} & 19.5\\
\hline
NeuroCRFs \cite{Do10} & 28.4 \\
\hline
Hidden-unit CRF  \cite{Maaten11} & 20.2\\
\hline
Deep learning  \cite{Hinton06b} & 8.57\\
\hline
Our method (without pretrain) & 27.1 \\ 
\hline
Our method  & {\bf 3.16} \\ 
\hline
\end{tabular}
}
\caption{The comparison on the CB513 dataset for protein secondary structure prediction task. It demonstrates that our method significantly outperforms other approaches. }
\label{tab:protein}
\end{table}

\begin{table*}[t!]
\centering
\resizebox{13.0cm}{!}{
\begin{tabular}{lccccccc}
\hline
\multirow{2}{*}{Methods} & \multicolumn{7}{c}{generalization error rate (\%)}  \\
    & 1000 & 2000 & 4000 & 5000 & 8000 & 10000 & 20000\\
\hline
Linear SVM & 12.22 & 12.0 & 10.6 & 9.33 & 8.96 & 8.71 & 8.27  \\ 
Deep learning \cite{Hinton06b} & 58.73 & 11.4 & 9.44 & 9.28 & 8.36 &  8.17 & 7.60 \\
Hidden CRF \cite{Maaten11}      & 10.2     & 8.74     &  7.59      & 7.45  & 7.02 & 6.79   &  6.29  \\
Our method                     & {\bf 9.79}  &  {\bf 7.97}  &   {\bf 6.66}    & {\bf 6.5}  & {\bf 6.26}  & {\bf 6.24}     & {\bf 6.01}  \\
\hline
\end{tabular}}
\caption{The experimental comparison on the treebank data set by varying the number of training data. It demonstrates that given few training data, our method is generalized well and more robust in the recognition task.  }
\label{tab:treebank}
\end{table*}

\section{Related work}
To predict structured output with deep learning is a challenge in machine learning \cite{Mnih11}. The difficulty of this problem is that the input and output data have non-determinated length, which may lead to an exponential number of possible configurations. Recently, a conditional RMBs is proposed for structured output prediction \cite{Mnih11}. Unfortunately, the model is shallow with only one hidden layer, and also cannot deal with large output configurations well. Typically, it either considers a small output space or uses semantic hashing in order to define and efficiently compute a small set of possible outputs to predict. More recently, SVM with deep learning \cite{Tang13} is proposed for classification tasks, such as object recognition. However, this model cannot handle the structured output, and thus cannot be used in sequential labeling problems. 

On the contrary, graphic models, such as hidden Markov model (HMM) and CRFs have been an popular method for segmentation and labeling time series data \cite{Lafferty01}. Over the last decade, many different approaches have been proposed to improve its performance on the sequential labeling problems. One trend is to extend the linear CRFs into the high-order graphical model, by exploiting more context information \cite{Koltun11,Cuong14}. However, the main weakness of those approaches is the time-consuming inference in the high-order graphical model. Another trend in the CRFs is to discover discriminative features to improve classification performance. One related work is a multilayer CRF (ML-CRF) \cite{Prabhavalkar10}. The system uses a multilayer perceptron (MLP), with one layer of hidden units, with a linear activation function for the output layer units and a sigmoid activation function for the hidden layer units. Similarly, hidden-unit CRFs \cite{Peng09,Maaten11} also assumes one-hidden layer for feature representation. The main idea of these two methods is similar to our approach here, in that we also transform the input to construct hidden features from the data so that these hidden units are discriminative in classification. But, unlike those systems, our model inherits the advantages of deep learning, and feature functions do not have any direct interpretation and are learned implicitly. Moreover, the deep features learned with large hidden units are powerful enough to represent the data, and generalize well in the classification tasks. As demonstrated by previous work, the performance of linear CRFs on a given task is strongly dependent on the feature representations \cite{Sutton06}; while deep learning \cite{Hinton06a} can learn representations that are helpful for classification. Thus, it is possible to unify these two methods into one framework. 
 
Our sequential labeling model with deep learning also bears some resemblance to approaches that train a deep network, and then train a linear CRF or Viterbi decoder on the output of the resulting network \cite{Do10,Mohamed09}. However, these methods differ from our approach in that (1) The deep feature learning and classification are separated steps; in other words, they use an unsupervised manner to train the weights from the data; (2) they do not learn all state-transition, data-dependent parameters and weights in the deep networks jointly. As a result, the top hidden units in these models may not discover latent distributed representations that are discriminative for classification. (3) Previous approaches \cite{LeCun98,Prabhavalkar10} does not take an online learning strategy to estimate model parameters. But we consider to update the weights with an online algorithm in our deep learning model, which can learn more useful representations \cite{Bengio12} to handle large scale dataset. Our work here inherits both advantages of CRFs and deep learning. Thus, our model can effectively handle structured prediction, and also learn discriminative features automatically for better sequential labeling under an unified framework. In addition, we use an online learning algorithm to update the model parameters, which is more robust than L-BFGS.

\section{Conclusions}
In this paper, we introduced a model to predict sequential labels with deep learning. Although deep learning can learn latent representations to improve classification, it cannot effectively handle structured output, such as sequential labeling. Hence, our approach unifies both advantages from linear-chain CRFs and deep learning, which can leverage both feature learning and context information for the classification and segmentation of time series. We use an simple but effective online learning method to update the model parameters, which has the potential to handle large-scale learning problem more effectively than widely used L-BFGS. In the experiments, we show that our model outperforms the current state of the art remarkably on a wide range of tasks.

%
%

\bibliography{rbmbib}
\bibliographystyle{ieee}

\end{document}